\documentclass[runningheads]{llncs}

\usepackage{eccv}

\usepackage{xcolor}
\newcommand{\rev}[1]{#1}

\usepackage{eccvabbrv}

\usepackage{graphicx}
\usepackage{booktabs}
\usepackage{amsmath,amssymb}
\usepackage{multirow}
\usepackage{makecell}
\usepackage{array} 
\usepackage{algorithm}
\usepackage{algpseudocode}
\usepackage{placeins}

\usepackage[accsupp]{axessibility}  

\usepackage{hyperref}
\hypersetup{
    colorlinks=true,
    urlcolor=eccvblue
}

\usepackage{orcidlink}

\begin{document}


\title{TaxoMIL: Taxonomy-Constrained Learning for Hierarchical Whole Slide Image Analysis} 

\titlerunning{TaxoMIL}

\author{Chaeyeon Lee\inst{1}\orcidlink{0009-0005-0608-9485} \and
Khang Nguyen Quoc\inst{1}\orcidlink{0000-0003-4927-4822} \and
 Jinsol Song\inst{1}\orcidlink{0009-0007-9190-2227} \and
 Yosep Chong\inst{2}\orcidlink{0000-0001-8615-3064} \and
 Kwangil Yim\inst{2}\orcidlink{0000-0001-8767-9033} \and
 Jin Tae Kwak\inst{1}\orcidlink{0000-0003-0287-4097}\thanks{Corresponding author.}}

\authorrunning{C. Lee et al.}

\institute{School of Electrical Engineering, Korea University, Seoul, Republic of Korea\\
\email{\{chaeyeonlee01,jkwak\}@korea.ac.kr} \and
Department of Hospital Pathology, The Catholic University of Korea College of Medicine, Seoul, Republic of Korea}

\maketitle
\begin{abstract}
Whole slide image (WSI) analysis is central to computational pathology, with multiple instance learning (MIL) emerging as the standard pipeline for slide-level diagnosis. However, conventional approaches formulate WSI diagnosis as a flat classification task over discrete labels, contradicting the inherently hierarchical, coarse-to-fine nature of clinical reasoning. Although recent hierarchical classifiers and vision--language models (VLMs) have sought to address this structural gap, they either fail to capture semantic continuity between related diagnoses or suffer from unconstrained text generation that produces taxonomic hallucinations and parent--child label violations. To address these limitations, we propose TaxoMIL, a taxonomy-constrained framework that reformulates WSI diagnosis as a multi-granularity text generation task. TaxoMIL utilizes a dual-head Transformer decoder to generate coarse- and fine-level diagnostic text, and introduces taxonomy-guided objectives that explicitly structure the label embedding space and strictly ground slide-level visual representations within the clinical taxonomy. Extensive experiments across three diverse WSI datasets demonstrate that TaxoMIL consistently outperforms state-of-the-art MIL classifiers and VLM-based generative methods, yielding accurate and hierarchy-aware diagnostic predictions. The code is released at \url{https://github.com/QuIIL/TaxoMIL}.

\end{abstract}

\section{Introduction}
\label{sec:intro}
\begin{figure}[!]
    \centering
    \includegraphics[width=1\linewidth]{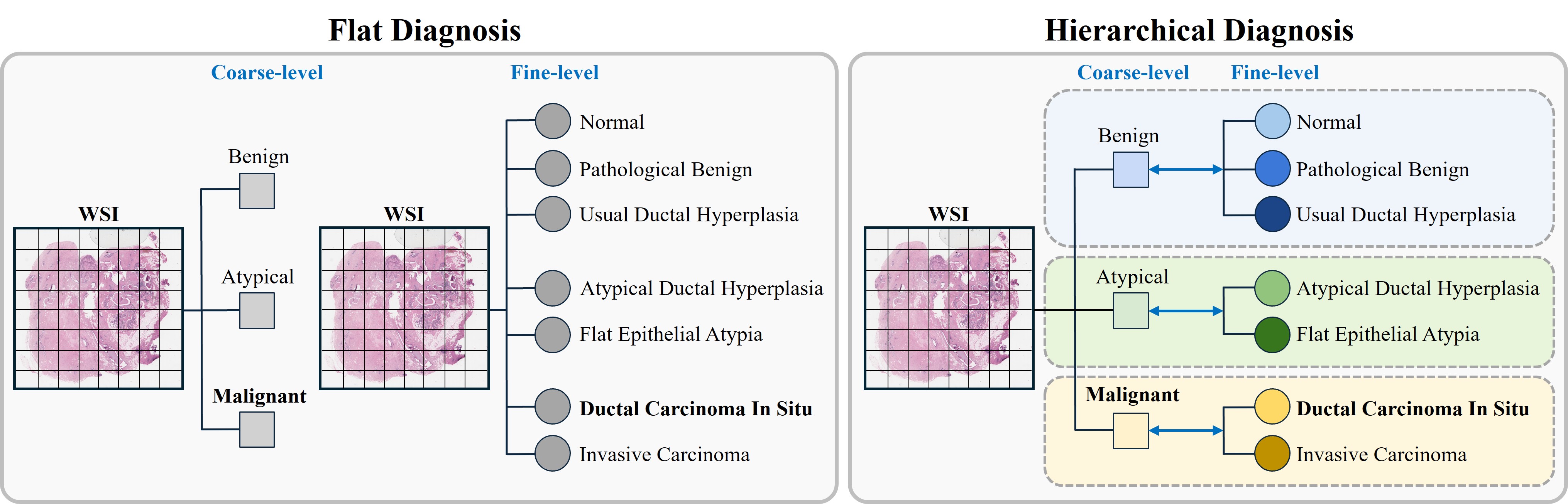}
    \caption{Flat vs. hierarchical WSI analysis. Flat classification treats labels as independent categories, ignoring clinical relationships. Hierarchical classification models the coarse-to-fine taxonomy, ensuring parent--child consistency across diagnostic levels.}
    \label{fig:flat_vs_hierarchical}
\end{figure}

In computational pathology, WSI analysis is a cornerstone for building decision-support diagnostic systems. The rapid progress in deep learning has accelerated the development of automated tools capable of analyzing and quantifying complex tissue and cellular characteristics in WSIs. However, due to the gigapixel-scale of WSIs, direct end-to-end optimization remains computationally intractable. Consequently, MIL methods have emerged as a practical pipeline, where a slide is decomposed into local patch-level instances, encoded into feature embeddings, and aggregated into a global slide-level representation for prediction. Throughout these standard frameworks, slide-level diagnosis has been overwhelmingly formulated as a flat, mutually exclusive classification task over discrete label indices, which contradicts the structured nature of clinical label space (Figure~\ref{fig:flat_vs_hierarchical}).

Clinical diagnosis is inherently hierarchical and often proceeds in a coarse-to-fine manner. A pathologist may first assign a broad diagnostic category (e.g., ``\textit{Malignant}'') and then refine the assessment into specific subtypes or grades (e.g., ``\textit{Ductal Carcinoma In Situ, Invasive Carcinoma}'') (Figure~\ref{fig:flat_vs_hierarchical}). 
Yet, when labels are treated as unrelated class indices, standard optimization objectives provide no explicit incentive to preserve coarse-to-fine (or parent--child) consistency or to encode semantic proximity among taxonomically related diagnoses. These gaps motivate moving beyond conventional flat classification toward formulations that natively capture the taxonomic relationships inherent in clinical reasoning.

Recent advances address this structural gap through two divergent paradigms: discrete hierarchical classifiers and unconstrained VLMs. Hierarchical classifiers \cite{jin2024hmil} incorporate label taxonomies via multi-branch architectures but still rely on flat, discrete class indices, failing to capture the semantic continuity between diagnostic labels. Conversely, VLM-based models \cite{slidechat, WSI-VQA} output clinically readable text that naturally accommodates varying granularities. However, robust diagnostic generation for WSIs remains challenging. Without explicit taxonomic constraints in the learned representation space, these generative models are prone to hallucinations, inconsistent diagnoses, and parent--child violations. Hence, neither paradigm simultaneously achieves structured reasoning, taxonomic faithfulness, and interpretable text generation.

To bridge this gap, we propose TaxoMIL, a taxonomy-constrained MIL framework that casts WSI diagnosis as hierarchical, taxonomy-consistent text generation. A standard MIL representation is modulated into coarse- and fine-level representations via a taxonomy-aware conditioning mechanism, and a dual-head Transformer decoder generates the corresponding diagnostic label text. To enforce clinical structure, taxonomy-guided objectives shape the label embedding space and align slide-level representations with taxonomy-consistent semantics, improving coarse-to-fine coherence and reducing hierarchical contradictions.

Our contributions are three-fold:
\begin{itemize}
\item We propose a taxonomy-constrained framework that reformulates WSI diagnosis as a multi-granularity text generation task to enable structured, hierarchy-aware clinical reasoning.
\item We introduce taxonomy-guided hierarchical objectives that explicitly structure the label embedding space and ground slide representations within the clinical taxonomy.
\item We evaluate on three WSI datasets (GastWSI, BRACS\cite{BRACS}, and PANDA\cite{PANDA}), demonstrating that TaxoMIL consistently outperforms state-of-the-art MIL classifiers and VLM-based methods.
\end{itemize}

\section{Related Work}
\label{sec:related_work}
\noindent\textbf{MIL for WSI Classification.}
WSIs are commonly modeled as bags of patch instances, where MIL aggregates instance features into a slide-level representation. Early approaches relied on permutation-invariant pooling (e.g., mean/max pooling), while attention-based MIL introduced instance weighting and became a standard backbone for WSI analysis \cite{abmil}. 
Subsequent works incorporated instance selection, multi-branch attention, or cluster-level constraints to handle intra-slide heterogeneity \cite{clam}, while Transformer-based aggregators modeled inter-instance interactions via self-attention \cite{transmil}. More recently, state space models \cite{s4mil, mambamil} offered linear-time aggregation for long-instance sequences.
In parallel, multimodal and language-guided MIL has been explored to inject semantic priors into WSI representations, such as ViLa-MIL\cite{vilamil}, which leverages vision--language pretraining to improve slide-level discrimination and interpretability.

\noindent\textbf{Hierarchical MIL Learning in Pathology.}
To better capture clinical taxonomies, several WSI methods explicitly incorporate label hierarchies. HMIL \cite{jin2024hmil} exploited label hierarchies at both instance and bag levels via a dual-branch design with class-wise attention, hierarchical alignment modules, and supervised contrastive learning. Beyond pathology, Chang et al.\ \cite{Flamingo} utilized top-down traversal over a coarse-to-fine label hierarchy, showing benefits from disentangling coarse- and fine-level feature learning.
While our approach conceptually aligns with these multi-granular, hierarchical formulations, we impose hierarchy directly on textual label embeddings and perform WSI-to-text generation rather than through multi-head classification over discrete IDs. 

\noindent\textbf{Vision--Language Modeling and Text Generation for Pathology.} 
VLMs combine a visual encoder with a pretrained language model to generate interpretable outputs, and recent pathology works extend this paradigm to slide- or region-level retrieval, question answering, and instruction-following assistants \cite{slidechat, WSI-VQA, nguyen2024camp}. 
WSI-VQA \cite{WSI-VQA} reframes diverse slide-level objectives into a generative visual question answering setting, enabling question-guided reasoning on WSIs. CAMP\cite{nguyen2024camp} similarly provides patch- and slide-level pathology classification that casts prediction as text generation, capturing task-specific knowledge via trainable adaptors. SlideChat\cite{slidechat} further moves toward instruction-following assistants, utilizing two-stage cross-domain alignment and visual instruction learning. These approaches can provide clinically readable outputs and reduce the semantic gap between model predictions and diagnostic terminology. Nevertheless, accurate and taxonomically coherent WSI diagnosis by generation remains challenging. 

\section{Methods}
\label{sec:methods}

\subsection{Problem Formulation}
In standard MIL practice, a WSI $I$ is decomposed into a disjoint set of patches: $\mathbf{P}=\{\mathbf{p}_i\}_{i=1}^{N_p}$, where $N_p$ denotes the number of patches. A feature extractor $\mathcal{E}(\cdot)$ encodes each patch into embedding space:  $\mathbf{x}_i = \mathcal{E}(\mathbf{p}_i)\in\mathbb{R}^{D}$, $\mathbf{X}=\{\mathbf{x}_i\}_{i=1}^{N_p}$, where $D$ is the embedding dimension. A MIL aggregator $\texttt{Agg}(\cdot)$ pools these patch-level embeddings into a slide-level embedding: $\mathbf{z}=\texttt{Agg}(\mathbf{X})\in\mathbb{R}^{D}$. For flat classification, $\mathbf{z}$ is fed to a linear classifier to predict a slide-level label $y \in \mathcal{Y}$, where $\mathcal{Y}$ denotes a set of classes. 

Clinical diagnosis, however, inherently follows a structured taxonomy. To formalize this, let $\mathcal{C}$ and $\mathcal{F}$ denote sets of coarse- and fine-level diagnostic text labels (i.e., natural language class names), respectively. The taxonomy mapping $\mathcal{H}: \mathcal{F} \rightarrow \mathcal{C}$ specifies the parent--child relationship, such that ground truth text labels satisfy $\mathcal{H}(\mathbf{y}^{(f)}) = \mathbf{y}^{(c)}, \mathbf{y}^{(c)} \in \mathcal{C}, \mathbf{y}^{(f)} \in \mathcal{F}$. 
The objective of TaxoMIL is to learn a mapping from the WSI space to taxonomically coherent text label pairs:
\begin{equation}
    (\hat{\mathbf{y}}^{(c)}, \hat{\mathbf{y}}^{(f)}) = f_{\theta}(I), \qquad \text{s.t.} \qquad \mathcal{H}(\hat{\mathbf{y}}^{(f)}) = \hat{\mathbf{y}}^{(c)}
\end{equation}
where $(\hat{\mathbf{y}}^{(c)}, \hat{\mathbf{y}}^{(f)})$ are the generated coarse- and fine-level diagnostic text strings, and $f_{\theta}$ represents the proposed taxonomy-constrained MIL framework. 

\subsection{Overview of TaxoMIL}
\begin{figure}
    \centering
    \includegraphics[width=1\linewidth]{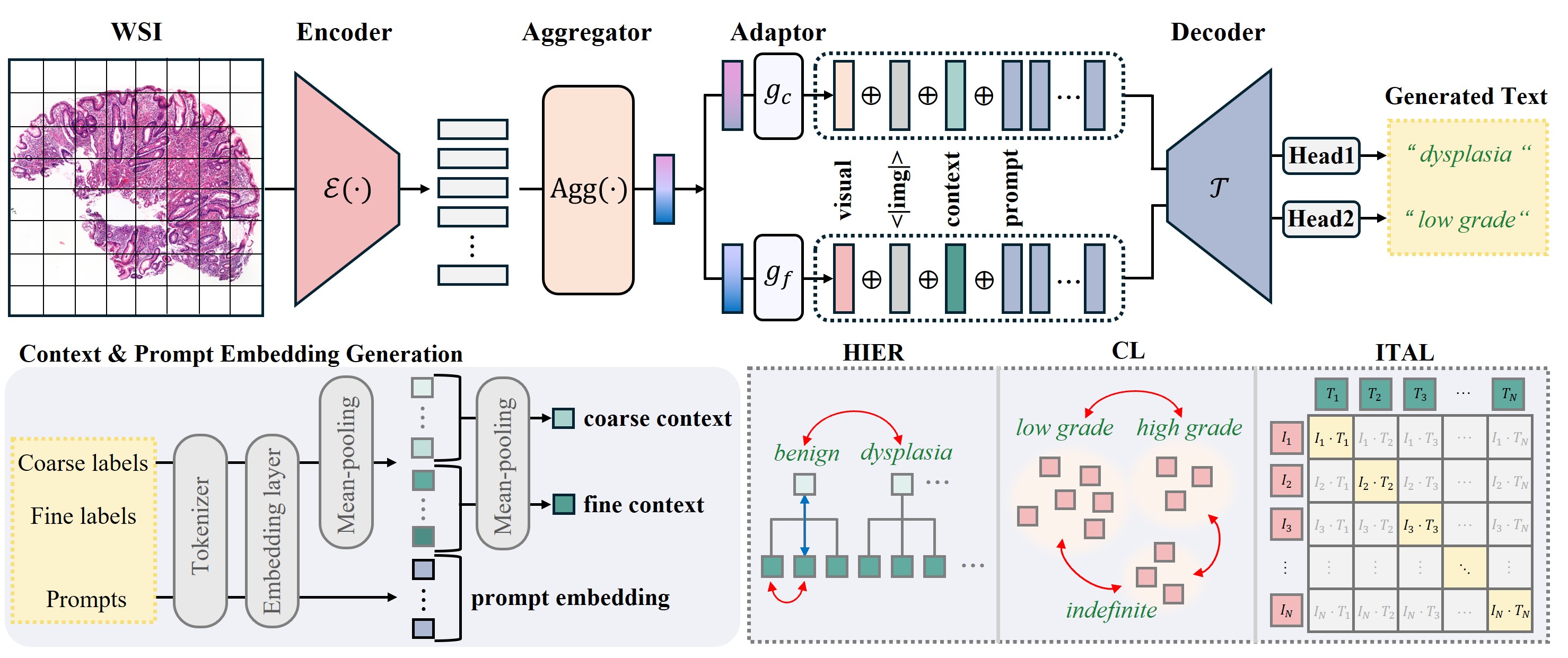}
    \caption{Overview of TaxoMIL. TaxoMIL consists of a MIL backbone, dual-branch multimodal conditioning module, and a dual-head text decoder. The model is optimized with label-text generation and supervised contrastive learning (CL) losses, augmented by taxonomy-guided label hierarchy (HIER) and image--text alignment (ITAL) objectives.}
    \label{fig:overview}
\end{figure}

TaxoMIL reframes WSI diagnosis as a taxonomy-constrained hierarchical text generation problem. As illustrated in Figure~\ref{fig:overview}, the framework comprises three major components: (1) a MIL backbone ($\mathcal{E}(\cdot)$, $\texttt{Agg}(\cdot)$) producing a global slide-level representation $\mathbf{z}\in\mathbb{R}^{D}$; (2) a dual-branch multimodal conditioning module $\mathcal{V}$ that transforms $\mathbf{z}$ into coarse- and fine-level visual representations conditioned on label context; and (3) a dual-head decoder $\mathcal{T}$ generating the corresponding diagnostic label text.
The algorithm of TaxoMIL is provided in Supp.~\ref{alg:taxomil_training}.

\subsection{Dual-branch Multimodal Conditioning Module}

Given the global slide-level representation $\mathbf{z}\in\mathbb{R}^{D}$ and semantic priors from the clinical taxonomy, the dual-branch conditioning module $\mathcal{V}$ produces two granularity-specific conditioning sequences for the dual-head decoder $\mathcal{T}$. This supports decoupled label-text generation and provides an interface for injecting taxonomy semantics.

\noindent\textbf{Embedding Decoupling.} 
The MIL backbone produces the global slide-level representation $\mathbf{z}\in\mathbb{R}^{D}$. To support decoupled hierarchical reasoning, we split $\mathbf{z}$ into two equal-sized embeddings along the feature dimension: $\mathbf{z} = [\mathbf{z}_c ; \mathbf{z}_f]$, where $\mathbf{z}_c$, $\mathbf{z}_f \in \mathbb{R}^{D/2}$ denote coarse- and fine-level embeddings, respectively. Each embedding is projected into the embedding space of the dual-head text decoder $\mathcal{T}$ via a dedicated lightweight adaptor: $\mathbf{v}^{(c)}=g_c(\mathbf{z}_c), \mathbf{v}^{(f)}=g_f(\mathbf{z}_f)$, where $\mathbf{v}^{(c)}, \mathbf{v}^{(f)} \in \mathbb{R}^H$, and $H$ is the embedding dimension of $\mathcal{T}$. Each adaptor is a linear projection followed by layer normalization and GELU activation. 


\noindent\textbf{Label Embedding Construction.} 
TaxoMIL uses label names not only as generation targets but also as semantic priors for taxonomy-constrained optimization. Recall that $\mathcal{C} = \{\mathbf{y}_i^{(c)}\}_{i=1}^{|\mathcal{C}|}$ and $\mathcal{F} = \{\mathbf{y}_i^{(f)}\}_{i=1}^{|\mathcal{F}|}$ denote the coarse- and fine-level text labels, respectively. Each text label is tokenized into a sequence of token IDs using $\texttt{TOK}$, the designated tokenizer of $\mathcal{T}$.
The corresponding token embeddings are retrieved and combined via mean-pooling to form base label-text representations $\{\mathbf{b}_i^{(c)}\}_{i=1}^{|\mathcal{C}|}$ and $\{\mathbf{b}_i^{(f)}\}_{i=1}^{|\mathcal{F}|}$. A shared learnable projection head $\phi(\cdot)$ is then applied to produce coarse- and fine-level label-text embeddings: $\{\mathbf{t}_i^{(c)}\}_{i=1}^{|\mathcal{C}|} = \{\phi (\mathbf{b}_i^{(c)})\}_{i=1}^{|\mathcal{C}|}$ and $\{\mathbf{t}_i^{(f)}\}_{i=1}^{|\mathcal{F}|} = \{\phi (\mathbf{b}_i^{(f)})\}_{i=1}^{|\mathcal{F}|}$. These embeddings serve as the semantic anchors for the taxonomy-constrained objectives.

\noindent\textbf{Conditioning Sequence Construction.}
To condition hierarchical label-text generation on both visual evidence and taxonomy semantics, we construct a short decoder prefix for each granularity by concatenating four components: (1) a granularity-specific visual token embedding ($\mathbf{v}^{(c)}$ or $\mathbf{v}^{(f)}$), (2) a learnable image token embedding ($\mathbf{e}_{\text{img}}$), (3) a granularity-specific label-context token embedding ($\mathbf{e}_{\text{ctx}}^{(c)}$ or $\mathbf{e}_{\text{ctx}}^{(f)}$), and (4) prompt token embeddings ($\mathbf{E}_{\text{prompt}}$). Formally, the coarse- and fine-level conditioning sequences are formed as follows:
\begin{equation}
\mathbf{E}^{(c)} = [\,\mathbf{v}^{(c)};\, \mathbf{e}_{\text{img}};\, \mathbf{e}^{(c)}_{\text{ctx}};\, \mathbf{E}_{\text{prompt}}\,], \qquad
\mathbf{E}^{(f)} = [\,\mathbf{v}^{(f)};\, \mathbf{e}_{\text{img}};\, \mathbf{e}^{(f)}_{\text{ctx}};\, \mathbf{E}_{\text{prompt}}\,].
\end{equation}
Here, $\mathbf{e}_{\text{img}} \in \mathbb{R}^{H}$ is a learnable token embedding for the special image token \texttt{<|img|>}, which serves as a fixed visual reference for decoding.
The label-context token embeddings $\mathbf{e}_{\text{ctx}}^{(c)}, \mathbf{e}_{\text{ctx}}^{(f)} \in \mathbb{R}^{H}$ provide global semantic priors over their respective label space, grounding the label-text generation in the clinical taxonomy. We compute these embeddings by averaging the corresponding base label embeddings across all classes at each granularity: $\mathbf{e}_{\text{ctx}}^{(c)}=\frac{1}{|\mathcal{C}|}\sum_{i=1}^{|\mathcal{C}|}\mathbf{b}_i^{(c)}$  and  $\mathbf{e}_{\text{ctx}}^{(f)}=\frac{1}{|\mathcal{F}|}\sum_{i=1}^{|\mathcal{F}|}\mathbf{b}_i^{(f)}$. 
The prompt token embeddings $\mathbf{E}_{\text{prompt}} \in \mathbb{R}^{L_p \times H}$ serve as a natural language instruction that steers the dual-head text decoder $\mathcal{T}$ toward generating short label-like outputs (i.e., class name phrases). The prompt is a fixed template string, tokenized and mapped to the embedding space of $\mathcal{T}$ with a prepended BOS token.
To reduce sensitivity to specific prompt phrasing, we adopt a template sampling strategy during training, randomly sampling from a curated set of 10 semantically equivalent but syntactically diverse templates. At inference, a fixed prompt is used for deterministic decoding. The prompt templates are listed in Supp.~\ref{app:prompts} and Table~\ref{tab:prompt_templates}.

\subsection{Dual-Head Text Decoder}
The dual-head text decoder $\mathcal{T}$ generates coarse- and fine-level diagnostic label text from the branch-specific conditioning sequences $\mathbf{E}^{(c)}$ and $\mathbf{E}^{(f)}$. To encourage consistent label text generation, we use a shared autoregressive backbone $\mathcal{T}_{shared}$ and attach two lightweight heads to perform granularity-specific generation. Specifically, given $\mathbf{E}^{(c)}$ and $\mathbf{E}^{(f)}$, $\mathcal{T}_{shared}$ produces hidden states: 
\begin{equation}
\mathbf{M}^{(c)}=\mathcal{T}_{shared}(\mathbf{E}^{(c)}) \in \mathbb{R}^{L^{(c)} \times H}, \qquad
\mathbf{M}^{(f)}=\mathcal{T}_{shared}(\mathbf{E}^{(f)}) \in \mathbb{R}^{L^{(f)} \times H}.
\end{equation}
where $L^{(c)}$ and $L^{(f)}$ denote the maximum token lengths for coarse- and fine-level predictions, respectively. 
Each hidden state is passed through a generation head (one Transformer decoder block) and a separate unembedding projection to produce token logits, from which the coarse- and fine-level label texts $\hat{\mathbf{y}}^{(c)}$ and $\hat{\mathbf{y}}^{(f)}$ are decoded autoregressively.

\subsection{Training Strategies}
TaxoMIL is trained end-to-end with a primary hierarchical label-text generation loss for both granularities, complemented by a supervised contrastive loss on visual embeddings and taxonomy-guided losses that (1) explicitly structure the label embedding space according to the clinical hierarchy and (2) ground visual embeddings in this taxonomy-structured label space.

\noindent\textbf{Overall Objective.}
The overall training objective is given by:
\begin{equation}
\mathcal{L} = w_{gen}\mathcal{L}_{GEN} + w_{hier}\mathcal{L}_{HIER} + w_{ital}\mathcal{L}_{ITAL} + w_{cl}\mathcal{L}_{CL},
\end{equation}
where $\mathcal{L}_{GEN}$ supervises hierarchical label-text generation, $\mathcal{L}_{HIER}$ structures the label embedding space according to the clinical taxonomy, $\mathcal{L}_{ITAL}$ grounds slide-level representations to the taxonomy-structured label space, and $\mathcal{L}_{CL}$ regularizes the visual embedding space. $w_{gen}$, $w_{hier}$, $w_{ital}$, and $w_{cl}$ denote the corresponding loss weights.

Let $\big\{(I_i, \mathbf{y}^{(c)}_i, \mathbf{y}^{(f)}_i)\big\}_{i=1}^{B}$ be a mini-batch of $B$ WSIs, where $I_i$ is a WSI and $y^{(c)}_i \in \mathcal{C}$ and $y^{(f)}_i \in \mathcal{F}$ denote the corresponding coarse- and fine-level text labels, respectively. For each WSI $I_i$, the MIL backbone ($\mathcal{E}(\cdot)$ and $\texttt{Agg}(\cdot)$) and dual-branch multimodal conditioning module $\mathcal{V}$ produce granularity-specific visual token embeddings $\mathbf{v}_i^{(c)}$ and $\mathbf{v}_i^{(f)}$ and coarse- and fine-level conditioning sequences $\mathbf{E}_i^{(c)}$ and $\mathbf{E}_i^{(f)}$. The dual-head text decoder $\mathcal{T}$ uses $\mathbf{E}_i^{(c)}$ and $\mathbf{E}_i^{(f)}$ to generate coarse- and fine-level text labels $\hat{\mathbf{y}}_i^{(c)}$ and $\hat{\mathbf{y}}_i^{(f)}$, respectively.

\noindent\textbf{Hierarchical Label-Text Generation Objective.}
For the label-text generation, we employ standard token-level cross-entropy for each granularity as $\mathcal{L}_{GEN}=\mathcal{L}_{GEN}^{(c)}+\mathcal{L}_{GEN}^{(f)}$, which serves as the primary loss.

\noindent\textbf{Label Hierarchical Loss.}
Given the projected coarse- and fine-level label-text embeddings $\{\mathbf{t}_i^{(c)}\}_{i=1}^{|\mathcal{C}|}$ and $\{\mathbf{t}_i^{(f)}\}_{i=1}^{|\mathcal{F}|}$, we enforce them to respect the predefined clinical taxonomy $\mathcal{H}$. Formally, we define the label hierarchy loss (HIER), which consists of (1) a hierarchy alignment loss $\mathcal{L}_{\text{HAL}}$ and (2) a sibling-margin loss $\mathcal{L}_{\text{SML}}$:
$\mathcal{L}_{HIER}=\mathcal{L}_{\text{HAL}}+\mathcal{L}_{\text{SML}}$.

\noindent\textbf{Hierarchy Alignment Loss.}
To enforce the clinical taxonomy in the label embedding space, the hierarchy alignment loss $\mathcal{L}_{\text{HAL}}$ is defined as:
\begin{equation}
\mathcal{L}_{\text{HAL}}
=
-\frac{1}{|\mathcal{C}|}
\sum_{k=1}^{|\mathcal{C}|}
\log
\left(
\frac{S_3(k)}{S_3(k)+S_1(k)+S_2(k)}
\right),
\end{equation}
where $S_1(k) = \sum_{k' \neq k} \exp\!\big( \mathbf{t}_k^{(c)}\cdot \mathbf{t}_{k'}^{(c)} \big)$, 
$S_2(k) = \frac{1}{n_k(n_k-1)} \sum_{\substack{i,j \in \mathcal{I}_k\\ i \neq j}}
\exp\!\big( \mathbf{t}_i^{(f)}\cdot \mathbf{t}_j^{(f)} \big)$,
$S_3(k) = \frac{1}{n_k} \sum_{i \in \mathcal{I}_k}
\exp\!\big( \mathbf{t}_k^{(c)}\cdot \mathbf{t}_i^{(f)} \big)$, and $\mathcal{I}_k = \{i \mid \mathcal{H}(i) = k\}$ defines the set of child fine-level indices for a coarse-level $k$, with $n_k = |\mathcal{I}_k|$.

This encourages that each coarse-level label embedding aligns with its descendant fine-level label embeddings ($S_3$) and remains separated from other coarse-level labels ($S_1$), while sibling fine-level label embeddings are separated from each other ($S_2$).

\noindent\textbf{Sibling Margin Loss.}
In $\mathcal{L}_{\text{HAL}}$, although $S_2$ implicitly encourages the relative separation of sibling fine label embeddings, these embeddings frequently collapse into highly clustered regions of embedding space. This likely occurs when the coarse--fine alignment term ($S_3$) dominates $\mathcal{L}_{\text{HAL}}$. To explicitly prevent sibling collapse, we formulate the sibling margin loss $\mathcal{L}_{\text{SML}}$ based on a hard Euclidean margin:
\begin{equation}
\mathcal{L}_{\text{SML}}
=
\frac{1}{|\mathcal{P}|}
\sum_{(i,j)\in\mathcal{P}}
\psi\!\left( m^2 - \|\mathbf{t}_i^{(f)}-\mathbf{t}_j^{(f)}\|_2^2 \right),
\end{equation}
where $m>0$ is a target margin, $\psi(\cdot)$ is a smooth hinge implemented with softplus, and the sibling pair set is defined as $\mathcal{P}=\cup_k \mathcal{P}_k$, with a parent-specific subset $\mathcal{P}_k=\{(i,j)\mid i<j,\; \mathcal{H}(i)=\mathcal{H}(j)=k\}$.



\noindent\textbf{Image--Text Alignment Loss.}
To ground the visual embeddings $\mathbf{v}_i^{(c)},\mathbf{v}_i^{(f)}$ in taxonomy-consistent label space, we align these embeddings to their corresponding label-text embeddings. For a branch $b \in \{c,f\}$, the image--text alignment loss (ITAL) is computed as:
\begin{equation}
\mathcal{L}_{ITAL}^{(b)}
=
-\sum_{i=1}^{B}
\log
\frac{
\exp\!\left( \mathbf{v}_i^{(b)\top} \mathbf{t}^{(b)}_{\mathbf{y}_i^{(b)}} / \tau \right)
}{
\sum_{j=1}^{N_b}
\exp\!\left( \mathbf{v}_i^{(b)\top} \mathbf{t}_j^{(b)} / \tau \right)
}
\end{equation}
where $\tau$ is a temperature and $N_b$ denotes total classes ($N_c=|\mathcal{C}|$ and $N_f=|\mathcal{F}|$). The total ITAL is obtained by summing branch-specific objectives: $\mathcal{L}_{ITAL} = \mathcal{L}_{ITAL}^{(c)} + \mathcal{L}_{ITAL}^{(f)}$.
This anchors visual embeddings to their ground-truth label semantics, forcing them to inherit the label-space hierarchy.

\noindent\textbf{Supervised Contrastive Loss.}
To further improve discriminability under weak slide-level supervision, we regularize the visual embeddings via supervised contrastive learning. For a branch $b \in \{c,f\}$, the supervised contrastive objective (CL) is formulated as:
\begin{equation}
\mathcal{L}_{CL}^{(b)} = - \sum_{i=1}^{B} \frac{1}{|\mathcal{G}^{(b)}(i)|} \sum_{g \in \mathcal{G}^{(b)}(i)}
\log 
\frac{\exp\!\left(\mathbf{v}_i^{(b)\top} \mathbf{v}^{(b)}_g / \tau \right)}
{\sum_{j \neq i} \exp\!\left(\mathbf{v}_i^{(b)\top} \mathbf{v}^{(b)}_j / \tau \right)}
\end{equation}
where $\mathcal{G}^{(b)}(i)$ denotes indices of positives for an anchor $i$. We compute $\mathcal{L}_{CL}^{(b)}$ independently for both branches and sum them to obtain the final CL loss: $\mathcal{L}_{CL} = \mathcal{L}_{CL}^{(c)} + \mathcal{L}_{CL}^{(f)}$.

\noindent\textbf{Cyclic Loss Scheduling.}
The overall objective includes four distinct loss terms. We fix $w_{gen}=1.0$ as generation is the primary task. Since $\mathcal{L}_{HIER}$ and $\mathcal{L}_{CL}$ enforce intra-modal structuring while $\mathcal{L}_{ITAL}$ promotes cross-modal alignment, optimizing all three jointly empirically induces instability. To mitigate this, we employ an alternating cyclic loss schedule that separates representation structuring from cross-modal alignment. 
Let $u=t/T \in [0,1]$ denote normalized training progress at epoch $t$ out of $T$ total epochs. We define a cyclic modulator $c(u)=\cos^2(\pi n_{\text{cycles}}u)$ to dynamically adjust the corresponding weights:
$w_{hier}(u)=w_{hier}^{\max}c(u)$, $w_{cl}(u)=w_{cl}^{\max}c(u)$, and
$w_{ital}(u)=w_{ital}^{\max}(1-c(u))$. In this manner, $\mathcal{L}_{HIER}$ and $\mathcal{L}_{CL}$ are emphasized when $\mathcal{L}_{ITAL}$ is down-weighted, and vice versa. 


\section{Experiments and Results}
\label{sec:experiments}
\subsection{Datasets and Label Hierarchies}
\label{subsec:dataset}
We evaluate TaxoMIL on three WSI datasets, each structured according to a two-level taxonomy with coarse-level labels and fine-level subtypes or grades.
GastWSI is a private dataset containing 7,228 gastric WSIs organized into 4 coarse-level and 23 fine-level categories. BRACS \cite{BRACS} and PANDA \cite{PANDA} are public breast and prostate datasets, respectively. BRACS includes 545 WSIs with 3 coarse-level and 7 fine-level labels.
PANDA comprises 10,614 WSIs annotated with 6 fine-level labels (normal and five ISUP grades). The fine-level labels are grouped into three coarse-level risk strata to construct a two-level hierarchy.
The details of the datasets and label hierarchies are summarized in Supp.~\ref{app:dataset} and Table~\ref{tab:data_hierarchy}.

\subsection{Evaluation Protocol}
\label{sec:eval_protocol}
For MIL-based baselines, predictions are evaluated using standard top-1 accuracy at both coarse and fine levels.
For text generation methods, including TaxoMIL and VLM-based baselines, we adopt a strict exact match evaluation protocol to ensure fair comparison. A generated text is correct only if the entire token sequence exactly matches the ground-truth label string, after decoding and removing special tokens.
Partial or prefix matches and semantically similar but non-identical outputs are strictly counted as incorrect to prevent diagnostic confusion in clinical settings.

\subsection{Baselines}
\label{subsec:baselines}
We compare TaxoMIL against a total of twelve baselines spanning MIL-based flat and hierarchical methods as well as VLM-based methods. 
\rev{All baselines were retrained from their official implementations under identical data splits, frozen UNI features, and evaluation protocols to ensure a fair comparison.}

\noindent\textbf{MIL-based Baselines.}
We employ five single-label MIL baselines (TransMIL\cite{transmil}, CLAM-SB\cite{clam}, ABMIL\cite{abmil}, MambaMIL\cite{mambamil}, and S4MIL\cite{s4mil}) and four hierarchical MIL baselines (HMIL\cite{jin2024hmil}, Chang et al.\cite{Flamingo}, ViLa-MIL\cite{vilamil}, and HiClass\cite{HiClass}). ViLa-MIL is originally a dual-magnification single-label classifier. We adapt its dual-scale design to a dual-granularity setting by training one branch on coarse-level labels and the other on fine-level labels.

\noindent\textbf{VLM-based Baselines.}
We include three VLM baselines (WSI-VQA\cite{WSI-VQA}, SlideChat\cite{slidechat}, and CAMP\cite{nguyen2024camp}), which perform text-based reasoning over WSIs.

\subsection{Implementation Details}
\label{subsec:implementation}
For all three WSI datasets, we extract patch embeddings using the frozen UNI\cite{UNI} foundation model as the feature extractor $\mathcal{E}$. 
The shared autoregressive text decoder backbone $\mathcal{T}_{shared}$ is initialized from the pretrained GPT-2 small (124M)\cite{GPT-2} with 12 decoder layers.
We use the standard GPT-2 tokenizer as $\texttt{TOK}$ for all text processing.
During training, we set the temperature $\tau=0.07$ for both $\mathcal{L}_{ITAL}$ and $\mathcal{L}_{CL}$, and we apply a margin $m=1.5$ for $\mathcal{L}_{SML}$. For the cyclic loss scheduling, we set the number of cycles to $n_{\text{cycles}}=3$ and define the maximum loss weights as $w_{hier}^{\max}=0.3$, $w_{ital}^{\max}=0.3$, and $w_{cl}^{\max}=0.1$ in all experiments.
Further implementation and training details are available in Supp.~\ref{app:implementation}.

\subsection{Main Results}
\label{subsec:main_results}

\begin{table}[!t]
\centering
\caption{Results of holistic evaluation on three datasets. 
}
\label{tab:main_results_joint}
\setlength{\tabcolsep}{3.5pt}
\scriptsize
\resizebox{\textwidth}{!}{%
\begin{tabular}{l|ccc|ccc|ccc}
\toprule
\multirow{2}{*}{Method} &
\multicolumn{3}{c|}{\textbf{GastWSI}} &
\multicolumn{3}{c|}{\textbf{BRACS}} &
\multicolumn{3}{c}{\textbf{PANDA}} \\
\cmidrule(lr){2-4}\cmidrule(lr){5-7}\cmidrule(lr){8-10}
& ACC (\%)& W-F1 & $\kappa$ & ACC (\%)& W-F1 & $\kappa$ & ACC (\%)& W-F1 & $\kappa$ \\
\midrule
TransMIL\cite{transmil}&  58.42   &    0.5684&    0.5362&    53.70 &    0.5150 &    0.4318&    49.72 &    0.4731 &    0.3486\\
CLAM-SB\cite{clam}&   62.42&   0.6033&   0.5793&   66.67 &   0.6660 &   0.5936&   49.06 &   0.4742 &   0.3475\\
ABMIL\cite{abmil}&    61.02&    0.5849&    0.5631&    \underline{68.52}&    \underline{0.7024}&    \underline{0.6159}&   49.06 &   0.4756 &   0.3471\\
MambaMIL\cite{mambamil}&   58.42&   0.5855&   0.5388&   62.96 &   0.6321 &   0.5489&   48.31 &   0.4828 &   0.3425\\
S4MIL\cite{s4mil}&   59.72&   0.5844&   0.5509&   61.11 &   0.6064 &   0.5116&   49.06 &   0.4672 &   0.3431\\
HMIL\cite{jin2024hmil}&   55.07&   0.5363&   0.4993&   61.11 &   0.6274&   0.5255&    
49.25 &    
0.4721 &    
0.3448\\
Chang et al.\cite{Flamingo} &   59.72&   0.6023&   0.5544&   64.81 &   0.6429 &   0.5604&   48.12 &   0.4545&   0.3292\\
ViLa-MIL\cite{vilamil}&   61.58&   0.5806&   0.5679&   \underline{68.52}&   0.6855 &   0.6130&    
47.84 &    
0.4733 &   0.3359\\
HiClass\cite{HiClass}      &   61.02&   0.5934&   0.5641&   61.11 &   0.6370 &   0.5333&    
50.09 &   0.4824 &   0.3552\\
WSI-VQA\cite{WSI-VQA}&   56.74&   0.5620&   0.4204&   64.81 &   0.6652 &   0.5696&    
\underline{50.28}&   \underline{0.4928}&   \underline{0.3648}\\
SlideChat\cite{slidechat}    &     54.79&     0.4969&     0.4885&   51.85&   0.4284&   0.3880&    
46.72 &   0.4331 &   0.3093\\
 CAMP\cite{nguyen2024camp}& \underline{63.26}& \underline{0.6044}& \underline{0.5874}& 66.67& 0.6726& 0.5895& 47.47& 0.4353&0.3145\\
TaxoMIL (\textbf{Ours})       &    \textbf{64.55}&  \textbf{0.6183}&   \textbf{0.6210}&   \textbf{75.78}&   \textbf{0.7662}&   \textbf{0.6974}&   \textbf{54.60}&   \textbf{0.5347}&   \textbf{0.4127}\\
\bottomrule
\end{tabular}%
}
\end{table}

\begin{table}[!t]
\centering
\caption{Results of coarse- and fine-level evaluations on three datasets. 
}
\label{tab:main_results_coarse_fine}
\setlength{\tabcolsep}{3.5pt}
\scriptsize
\resizebox{\textwidth}{!}{%
\begin{tabular}{ll|ccc|ccc|ccc}
\toprule
& \multirow{2}{*}{Method} &
\multicolumn{3}{c|}{\textbf{GastWSI}} &
\multicolumn{3}{c|}{\textbf{BRACS}} &
\multicolumn{3}{c}{\textbf{PANDA}} \\
\cmidrule(lr){3-5}\cmidrule(lr){6-8}\cmidrule(lr){9-11}
&  & ACC (\%)& W-F1 & $\kappa$ & ACC (\%)& W-F1 & $\kappa$ & ACC (\%)& W-F1 & $\kappa$ \\
\midrule
\multirow{13}{*}{\rotatebox[origin=c]{90}{Coarse-level}} & TransMIL\cite{transmil}     &      83.91 &     0.8387 &     0.7848&     83.33 &     0.8435 &     0.7276&     69.61 &     0.6628 &     0.4409\\
& CLAM-SB\cite{clam}      &   86.42 &   0.8646 &   0.8180&   85.19&   0.8444 &   0.7410&   69.61 &   0.6604 &   0.4424\\
& ABMIL\cite{abmil}        &    86.60&    0.8661&    0.8205&    87.04 &    0.8676 &    0.7771&   69.32 &   0.6691 &   0.4247\\
& MambaMIL\cite{mambamil}     &   83.53 &   0.8336 &   0.7783&   83.33 &   0.8490 &   0.7370&   68.95 &   0.6605 &   0.4359\\
& S4MIL\cite{s4mil}        &   \underline{86.98}&   \underline{0.8691}&   \underline{0.8254}&   \underline{88.89}&   0.8878 &   0.8099&   \underline{70.36}&   \underline{0.6749}&   \underline{0.4609}\\
& HMIL\cite{jin2024hmil}         &   84.19&   0.8417&   0.7877&   85.19&   0.8495&   0.7435&    69.89&    0.6675&    0.4485\\
& Chang et al.\cite{Flamingo} &   86.42&   0.8656&   0.8174&   \underline{88.89}&   \underline{0.8898}&   \underline{0.8121}&   68.67&   0.6516&   0.4247\\
& ViLa-MIL\cite{vilamil}     &   85.95&   0.8578&   0.8120&   87.04&   0.8776&   0.7862 &    68.39&    0.6558&   0.4291\\
& HiClass\cite{HiClass}      &   85.86&   0.8605&   0.8097&   81.48 &   0.8378&   0.7072 &    69.42 &   0.6621&   0.4434\\
& WSI-VQA\cite{WSI-VQA}     &   83.44&   0.8336&   0.7779&   81.48&   0.8236&   0.6917&    70.17 &   0.6745 &   0.4604\\
& SlideChat\cite{slidechat}    &       85.49&       0.8546&       0.8054&  75.93&   0.7393&   0.5720&    68.95 &   0.6579 &   0.4334\\
&  CAMP\cite{nguyen2024camp}& 86.79& 0.8680& 0.8226& 87.04& 0.8776& 0.7862& 67.73& 0.6357&0.4055\\
& TaxoMIL (\textbf{Ours})       &    \textbf{88.56}&  \textbf{0.8856}&   \textbf{0.8466}&   \textbf{90.74}&   \textbf{0.9089}&   \textbf{0.8438}&   \textbf{71.29}&   \textbf{0.6890}&   \textbf{0.4787}\\
 \midrule
\multirow{13}{*}{\rotatebox[origin=c]{90}{Fine-level}} &  TransMIL\cite{transmil}     &  60.09 &  0.5722 &  0.5526&  53.70 &  0.5049 &  0.4275&   50.94 &  0.4790 & 0.3608\\
&  CLAM-SB\cite{clam}      & \underline{63.53}& 0.6015& \underline{0.5896}& \underline{70.37}& 0.6826 & 0.6351& 49.81 & 0.4781 &0.3529\\
&  ABMIL\cite{abmil}        & 62.51& 0.5816& 0.5771& \underline{70.37}& \underline{0.7147}& \underline{0.6370}& \underline{51.13}& 0.4867 &0.3487\\
&  MambaMIL\cite{mambamil}     & 61.21 & 0.5933& 0.5666& 66.67 & 0.6662 & 0.5909& 50.19& 0.4936&0.3600\\
&  S4MIL\cite{s4mil}        & 60.28 & 0.5783 & 0.5549& 61.11 & 0.5909 & 0.5048& 50.38 & 0.4751 &0.3556\\
&  HMIL\cite{jin2024hmil}         & 57.40& 0.5257& 0.5185& 61.11& 0.5971& 0.5120& 50.66& 0.4775&0.3573\\
&  Chang et al.\cite{Flamingo} & 62.05& 0.5966& 0.5760& 64.81& 0.6367& 0.5589& 50.00& 0.4688&0.3487\\
&  ViLa-MIL\cite{vilamil}     & 62.33& 0.5755& 0.5740& 68.52& 0.6831& 0.6123& 49.25& 0.4790&0.3486\\
&  HiClass\cite{HiClass}      & 62.79& 0.5937& 0.5812& 66.67 & 0.6784 & 0.5947& \underline{51.13}& 0.4876 &0.3651\\
&  WSI-VQA\cite{WSI-VQA}     & 58.23& 0.5667& 0.5342& 64.81& 0.6492& 0.5656& 50.84 & \underline{0.4968}&\underline{0.3702}\\
&  SlideChat\cite{slidechat}    &   56.93&   0.5015&   0.5089& 51.85& 0.4150& 0.3727& 47.56 & 0.4396 &0.3171\\
&  CAMP\cite{nguyen2024camp}& 63.26& \underline{0.6044}& 0.5874& 66.67& 0.6726& 0.5895& 47.47& 0.4353&0.3145\\
&  TaxoMIL (\textbf{Ours})       & \textbf{65.86}& \textbf{0.6183}& \textbf{0.6154}& \textbf{77.78}& \textbf{0.7632}& \textbf{0.7178}& \textbf{55.07}& \textbf{0.5358}&\textbf{0.4172}\\
\bottomrule
\end{tabular}%
}
\end{table}

We assess TaxoMIL against state-of-the-art MIL classifiers and generative VLMs across three evaluation scenarios: (1) Holistic evaluation (exact match required at both coarse and fine levels), (2) Coarse-level evaluation, and (3) Fine-level evaluation. 
We report accuracy (ACC), weighted F1 (W-F1), and Cohen's kappa ($\kappa$) to capture overall correctness, robustness to class imbalance, and agreement beyond chance, respectively. 
\rev{The computational efficiency of TaxoMIL and competing baselines is provided in Table~S5.}

\noindent\textbf{Holistic Evaluation.}
Table~\ref{tab:main_results_joint} reports holistic performance across all three datasets, serving as a strict benchmark for clinical usability, as it requires both coarse- and fine-level predictions to be correct. 
TaxoMIL achieved the best holistic performance on all three datasets by consistent margins in all evaluation metrics. Compared to the strongest per-dataset baseline, TaxoMIL yielded substantial gains by +1.29\% ACC, +0.0139 W-F1, and +0.0336 $\kappa$ on GastWSI (vs. CAMP), +7.26\% ACC, +0.0638 W-F1, and +0.0815 $\kappa$ on BRACS (vs. ABMIL), and +4.32\% ACC, +0.0419 W-F1, and +0.0479 $\kappa$ on PANDA (vs. WSI-VQA). Notably, none of the baselines demonstrated consistent performance across all three datasets. While CAMP, ABMIL, and WSI-VQA exhibited competitive results on a single dataset, their performance degraded substantially on the remaining two.
Overall, these results indicate stronger end-to-end coarse-to-fine consistency of TaxoMIL. 

\noindent\textbf{Coarse-level Evaluation.}
As shown in Table~\ref{tab:main_results_coarse_fine}, the overall performance of all models increased substantially at the coarse-level compared to the holistic evaluation, reflecting the reduced complexity of broader diagnostic categories. Nevertheless, TaxoMIL consistently outperformed both MIL and VLM baselines. Specifically, TaxoMIL provided performance gains of 1.58\% to 5.12\% ACC for GastWSI, 1.85\% to 14.81\% ACC for BRACS, and 0.93\% to 3.56\% ACC for PANDA, relative to baselines. Consistent improvements were also observed for W-F1 and $\kappa$, further validating the reliability of TaxoMIL.

\noindent\textbf{Fine-level Evaluation.}
At the fine-level, TaxoMIL maintained its superior performance (Table~\ref{tab:main_results_coarse_fine}), demonstrating its ability to resolve the subtle morphological ambiguities. Relative to the coarse-level evaluation, TaxoMIL obtained even larger performance margins over baselines; for instance, 2.33\% to 8.93\% ACC for GastWSI, 7.41\% to 25.93\% ACC for BRACS, and 3.94\% to 7.60\% ACC for PANDA. These consistently larger improvements at the fine-level suggest that taxonomy-constrained text generation and the hierarchy-guided objectives particularly benefit fine-level discrimination, where inter-class boundaries are more ambiguous and precise hierarchy-aware reasoning is most critical.

\subsection{Ablation Studies}
\label{subsec:ablation}

We conducted systematic ablation studies to assess the impact of the proposed training strategies and decoder prefix design.

\begin{table*}[!t]
\centering
\caption{Ablation study of training strategies on three datasets. 
}
\label{tab:ablation_progressive}
\setlength{\tabcolsep}{3.0pt}
\scriptsize
\resizebox{\textwidth}{!}{%
\begin{tabular}{llcccc|ccc|ccc|ccc}
\toprule

& \multicolumn{5}{c|}{Components}&
\multicolumn{3}{c|}{\textbf{GastWSI}} &
\multicolumn{3}{c|}{\textbf{BRACS}} &
\multicolumn{3}{c}{\textbf{PANDA}} \\
\cmidrule(lr){2-6}\cmidrule(lr){7-9}\cmidrule(lr){10-12}\cmidrule(lr){13-15}
  &  $\mathcal{L}_{GEN}$ & $\mathcal{L}_{CL}$ & $\mathcal{L}_{HIER}$ & $\mathcal{L}_{ITAL}$ & Cyclic & ACC (\%) & W-F1 & $\kappa$ & ACC (\%) & W-F1 & $\kappa$ & ACC (\%) & W-F1 & $\kappa$ \\
\midrule

\multirow{4}{*}{\rotatebox[origin=c]{90}{\centering Holistic}}  & \checkmark &&        &        &        &   61.58 &   0.6014 &   0.5714&   66.67 &   0.6409 &   0.5850&   49.53 &   0.4895 &   0.3585\\
  & \checkmark &\checkmark &     &        &        &  62.60 &  0.6053 &  0.5821&  62.96 &  0.5558 &  0.5246&  51.97&  0.5124&  0.3855\\
  & \checkmark &\checkmark & \checkmark & \checkmark &  &   63.07 &   0.6141 &   0.5879&   72.22 &   0.7205 &   0.6521&   51.41&   0.5092&   0.3793\\
  & \checkmark &\checkmark & \checkmark & \checkmark & \checkmark &  \textbf{64.55}&  \textbf{0.6183}&  \textbf{0.6210}&  \textbf{75.78}&  \textbf{0.7662}&  \textbf{0.6974}&  \textbf{54.60}&  \textbf{0.5347}&  \textbf{0.4127}\\
\midrule

\multirow{4}{*}{\rotatebox[origin=c]{90}{\parbox[c]{1cm}{\centering Coarse-level}}}  & \checkmark &&        &        &        &   87.35 &    0.8738 &    0.8304&   85.19 &   0.8620 &   0.7578&   70.54 &   0.6798 &   0.4659\\
  & \checkmark &\checkmark &     &        &        &  88.00 &  0.8800 &  0.8391&  90.74 &  0.9089&  0.8438&  71.20&  0.6837&  0.4738\\
  & \checkmark &\checkmark & \checkmark & \checkmark &  &  88.19 &  0.8819 &  0.8418&  88.89 &  0.8930 &  0.8146&  70.36&  0.6749&  0.4609\\
  & \checkmark &\checkmark & \checkmark & \checkmark & \checkmark &  \textbf{88.56}&  \textbf{0.8856}&  \textbf{0.8466}&  \textbf{90.74}&  \textbf{0.9089}&  \textbf{0.8438}&  \textbf{71.29}&  \textbf{0.6890}&  \textbf{0.4787}\\
\midrule

\multirow{4}{*}{\rotatebox[origin=c]{90}{\parbox[c]{1cm}{\centering Fine-level}}} & \checkmark &&        &        &        &   63.53 &   0.5974 &   0.5900&   70.37 &   0.6821 &   0.6278&   51.88 &   0.5015 &   0.3794\\
  & \checkmark &\checkmark &     &        &        &   63.63 &  0.5946 &  0.5905&  62.96&  0.5489&  0.5209&  52.53&  0.5126&  0.3897\\
  & \checkmark &\checkmark & \checkmark & \checkmark &  &  64.47 &  0.6093 &  0.6005&  74.07 &  0.7214 &  0.6727&  52.63&  0.5170&  0.3922\\
  & \checkmark &\checkmark & \checkmark & \checkmark & \checkmark &  \textbf{65.86}&  \textbf{0.6183}&  \textbf{0.6154}&  \textbf{77.78}&  \textbf{0.7632}&  \textbf{0.7178}&  \textbf{55.07}&  \textbf{0.5358}&  \textbf{0.4172}\\
\bottomrule
\end{tabular}%
}
\end{table*}

\begin{table}[!t]
\centering
\caption{Ablation study of the decoder conditioning sequence on three datasets. 
}
\label{tab:ablation_tokens}
\setlength{\tabcolsep}{3.2pt}
\scriptsize
\resizebox{\textwidth}{!}{%
\begin{tabular}{ll|ccc|ccc|ccc}
\toprule
& \multirow{2}{*}{Model Variant} &
\multicolumn{3}{c|}{\textbf{GastWSI}} &
\multicolumn{3}{c|}{\textbf{BRACS}} &
\multicolumn{3}{c}{\textbf{PANDA}} \\
\cmidrule(lr){3-5}\cmidrule(lr){6-8}\cmidrule(lr){9-11}
&
& ACC (\%) & W-F1 & $\kappa$ & ACC (\%) & W-F1 & $\kappa$ & ACC (\%) & W-F1 & $\kappa$ \\
\midrule

\multirow{3}{*}{\rotatebox[origin=c]{90}{\parbox[c]{0.8cm}{\centering \tiny Holistic}}} & \textbf{Full}                 &   \textbf{64.55}&   \textbf{0.6183}&   \textbf{0.6210}&   \textbf{75.78} &   \textbf{0.7662} &   \textbf{0.6974}&   \textbf{54.60} &   \textbf{0.5347} &   \textbf{0.4127}\\
& w/o $\mathbf{e}_{\text{ctx}}$    &    62.51&    0.6030&    0.5803&    70.37&    0.7006&    0.6285&    50.56 &    0.4915 &    0.3647\\
& w/o \texttt{<|img|>} &   62.05&   0.6072&   0.5766&   70.37&   0.6647&   0.6295&   53.10 &   0.5232 &   0.4065\\
\midrule

\multirow{3}{*}{\rotatebox[origin=c]{90}{\parbox[c]{0.8cm}{\centering \tiny Coarse-level}}} & \textbf{Full}                 &   \textbf{88.56}&   \textbf{0.8856}&    \textbf{0.8466}&   \textbf{90.74}&   \textbf{0.9089}&   \textbf{0.8438}&   \textbf{71.29} &   \textbf{0.6890} &   \textbf{0.4787}\\
& w/o $\mathbf{e}_{\text{ctx}}$    &  87.63 &  0.8763 &  0.8342&  87.04&  0.8776&  0.7862&  69.51 &  0.6600 &  0.4360\\
& w/o \texttt{<|img|>} &  87.72 &  0.8774 &  0.8355&  88.89 &  0.8930 &  0.8146&  70.92 &  0.6831 &  0.4718\\
\midrule

\multirow{3}{*}{\rotatebox[origin=c]{90}{\parbox[c]{0.8cm}{\centering \tiny Fine-level}}} & \textbf{Full}                 &   \textbf{65.86}&   \textbf{0.6183}&   \textbf{0.6154}&   \textbf{77.78}&   \textbf{0.7632}&   \textbf{0.7178}&   \textbf{55.07} &   \textbf{0.5358} &   \textbf{0.4172}\\
& w/o $\mathbf{e}_{\text{ctx}}$    &  64.47 &  0.6089 &  0.6000&  74.07 &  0.7130 &  0.6687&  51.22 &  0.4943 &  0.3700\\
& w/o \texttt{<|img|>} &  64.19&  0.6112 &  0.5979&  72.22 &  0.6929 &  0.6509&  53.85 &  0.5261 &  0.4065\\

\bottomrule
\end{tabular}%
}
\end{table}

\noindent\textbf{Effect of Training Strategies.}
Table~\ref{tab:ablation_progressive} presents a progressive ablation study, incrementally adding each objective to a base model trained with the text generation loss ($\mathcal{L}_{GEN}$). Adding the supervised contrastive learning loss ($\mathcal{L}_{CL}$) yielded modest but consistent gains across datasets, likely due to improved discriminability among visually similar slides. Introducing the label hierarchical loss ($\mathcal{L}_{HIER}$) and image--text alignment loss ($\mathcal{L}_{ITAL}$) produced the largest improvements, concentrated at the fine-level and holistic evaluations. On BRACS, fine-level and holistic ACCs improved by approximately +10.00\%, demonstrating the efficacy of hierarchy-guided objectives for fine-level discrimination. Notably, adding $\mathcal{L}_{CL}$ alone on BRACS achieved stronger coarse-level performance but degraded fine-level results, suggesting that $\mathcal{L}_{CL}$ without hierarchical constraints can lead to over-clustering of coarse categories at the expense of sibling discrimination. Finally, the inclusion of cyclic scheduling (full model) delivered the highest performance across all datasets. These results indicate that each component contributes complementary benefits for robust coarse-to-fine learning.


\noindent\textbf{Effect of Decoder Design.}
Table~\ref{tab:ablation_tokens} examines the impact of the decoder conditioning sequence by removing the learnable image token ($\texttt{<|img|>}$) or label-context token ($\mathbf{e}_{\text{ctx}}$). Removing either component consistently degraded performance across all granularities. For example, under holistic evaluation, the exclusion of $\texttt{<|img|>}$ decreased ACC by 2.50\%, 5.41\%, and 1.50\% on GastWSI, BRACS, and PANDA, respectively.
Similarly, omitting $\mathbf{e}_{\text{ctx}}$ yielded comparable drops of 2.04\%, 5.41\%, and 4.04\% on the same datasets. These consistent drops highlight their roles in structural grounding. Specifically, $\texttt{<|img|>}$ helps to disentangle visual conditioning from prompt tokens, while $\mathbf{e}_{\text{ctx}}$ provides a global semantic prior. 
\rev{We further examined two architectural variants: replacing the text decoder with classification heads (TaxoMIL-Cls) and varying the decoder backbone (Supp.~\ref{app:additional_analyses} and Table~\ref{tab:classifier_decoder}).
The generative formulation consistently improves holistic performance over its classification counterpart, and the lightweight GPT-2 small backbone yields the best accuracy--efficiency trade-off, indicating that the gains stem primarily from taxonomy-aware representation learning rather than decoder capacity.}

\subsection{Qualitative Assessments}
\begin{figure}[t]
    \centering
    \includegraphics[width=1\linewidth]{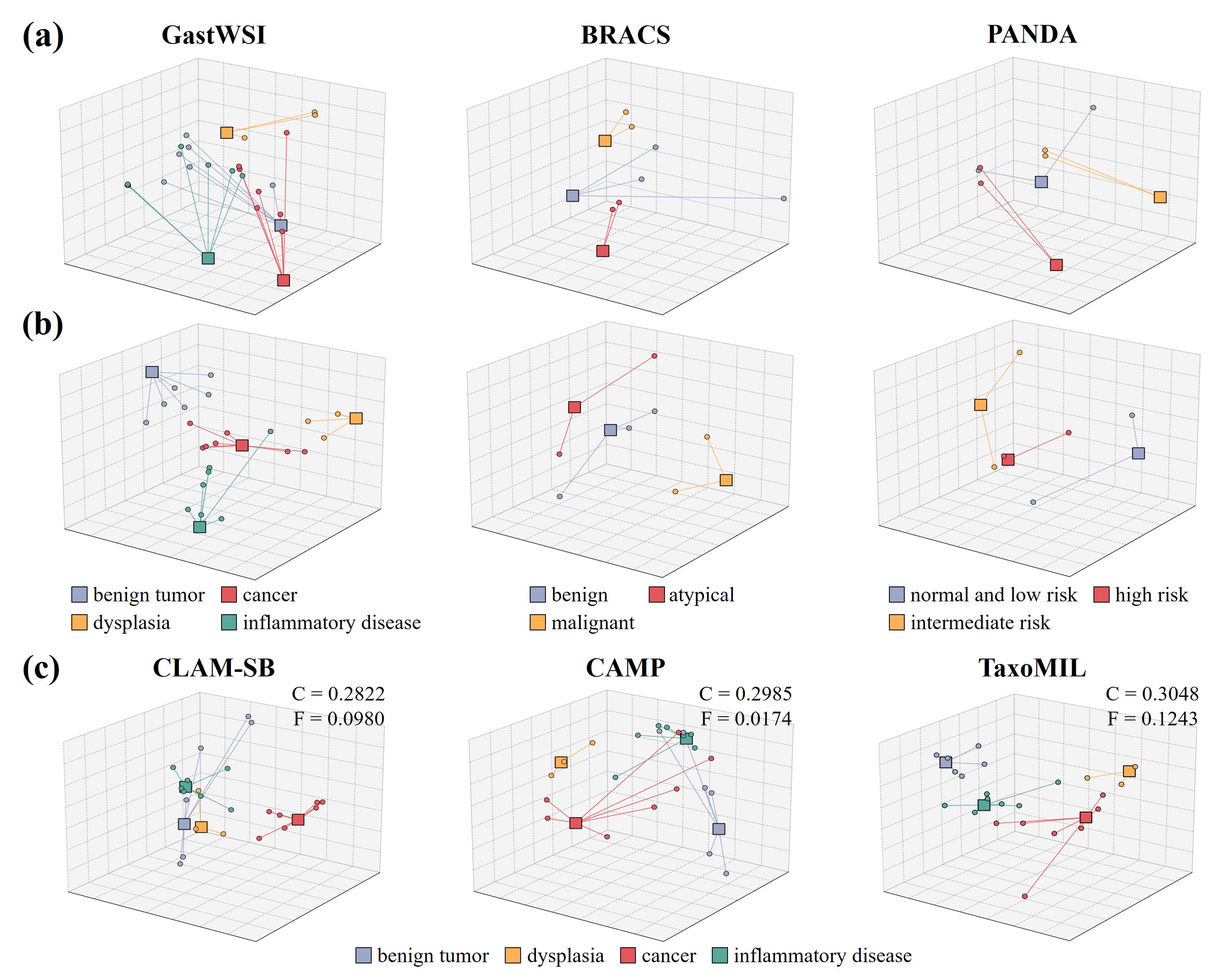}
    \caption{\textbf{Visualization of label-text and image embeddings.} MDS projections compare \textbf{(a)} the raw label-text embeddings (from a pretrained GPT-2), \textbf{(b)} the taxonomy-constrained label-text embeddings learned by TaxoMIL, and \textbf{(c)} image embedding centroids of coarse- and fine-level categories on the GastWSI dataset. Rectangles and circles represent coarse- and fine-level labels, respectively. C and F denote silhouette scores for coarse- and fine-level labels, respectively, summarizing cluster separation within each group.}
    \label{fig:label_geometry}
\end{figure}
\noindent\textbf{Label Embedding Geometry.}
\label{subsec:label_geometry} 
To investigate the effectiveness of the proposed hierarchy-guided objectives, we visualize the label-text embeddings $\{\mathbf{t}_k^{(c)}\} \cup \{\mathbf{t}_i^{(f)}\}$ by projecting them into a lower-dimensional manifold using multidimensional scaling (MDS). 
Figure~\ref{fig:label_geometry}\textbf{a} and \textbf{b} compare the resulting manifolds before and after taxonomy-constrained training. Prior to training, fine-level labels intermingle across coarse-level categories, and sibling fine-level labels frequently collapse, demonstrating the lack of explicit taxonomic structure in the raw pretrained embedding space. 
After training, we observe a markedly more taxonomy-aligned organization: (1) coarse-level labels gravitate toward the centroids of their child fine-level labels, serving as structural anchors for the hierarchy, (2) unrelated coarse-level categories become better separated, reducing high-level diagnostic confusion, and (3) sibling fine-level labels become more dispersed, facilitating fine-level discrimination. These observations closely match the design principles of the proposed hierarchy-guided objectives.

\noindent\textbf{Image Embedding Centroid Geometry.}
To further examine the organization of the learned image representations, we apply the same MDS technique to project slide-level embeddings from the GastWSI dataset. 
Figure \ref{fig:label_geometry}\textbf{c} visualizes the class centroids for both granularities, computed by averaging slide-level embeddings from the same class, and compares TaxoMIL against a representative MIL model (CLAM-SB) and a representative VLM model (CAMP). 
Relative to these baselines, TaxoMIL exhibits clearer separation among coarse-level categories while producing a highly structured fine-class layout within each coarse category, which is consistent with the label-text geometry. In contrast, CLAM-SB and CAMP show intermingled and overlapping class structures, indicating weaker hierarchical organization. These observations align with the improved coarse- and fine-level silhouette scores achieved by TaxoMIL (Figure \ref{fig:label_geometry}\textbf{c}).

\section{Limitations and Future Work}
\rev{
TaxoMIL assumes a fixed taxonomy defined by label-text embeddings and parent--child relations. Although it can be extended to deeper hierarchies via level-wise alignment and sibling-margin constraints, adapting to evolving taxonomies would require dynamic taxonomy construction, which we leave for future work. TaxoMIL also reduces but does not fully eliminate parent--child violations, as shown by the PCVR analysis in Supp.~\ref{app:additional_analyses} and Table~\ref{tab:pcvr}, and incurs additional parameter and memory overhead from the autoregressive decoder, as discussed in Supp.~\ref{app:computational_efficiency} and Table~\ref{tab:efficiency}.}

\section{Conclusion}
\label{sec:conclusion}
We present TaxoMIL, a taxonomy-constrained WSI-to-text label generation framework for hierarchical pathology diagnosis. Built on a standard MIL backbone for slide-level representations, TaxoMIL employs a dual-head autoregressive decoder to jointly generate coarse-level categories and corresponding fine-level subtypes within a unified generative pipeline. To enforce clinical structure, hierarchy-guided objectives organize the label embedding space according to the clinical taxonomy, align slide-level representations with taxonomy-consistent semantics, and improve parent--child coherence while reducing hierarchical contradictions. Across three WSI benchmarks, TaxoMIL yields more accurate and hierarchically consistent diagnostic predictions, outperforming state-of-the-art MIL classifiers and VLM methods. These results highlight the benefit of embedding clinical taxonomic structure directly into the learning objective, supporting interpretable and clinically meaningful diagnostic reasoning.

\section*{Acknowledgements}
This work was supported by a grant of the National Research Foundation of Korea (NRF) (No. RS-2025-00558322 and RS-2024-00397293).

%
%
\bibliographystyle{splncs04}
\bibliography{main}

\clearpage 
\setcounter{page}{1}

\appendix
\begin{center}
{\Large\bfseries TaxoMIL: Taxonomy-Constrained Learning for\\[0.3em]
Hierarchical Whole Slide Image Analysis}\\[0.65em]

{\large\bfseries -- Supplementary Material --}
\end{center}


\setcounter{table}{0}
\renewcommand{\thetable}{S\arabic{table}}

\section{Algorithm}
\label{alg:taxomil_training}

\begin{algorithm}[h]
\caption{TaxoMIL Training Process}
\small
\begin{algorithmic}[1]
\State \textit{\# Phase 1: Label-text initialization}
\State $(b^c,b^f) \gets \Call{LabelTextEmbeddings}{\mathcal{C},\mathcal{F}}$ \Comment{mean-pooled token embeddings}
\State $(t^c,t^f) \gets \Call{Projection}{\mathcal{C},\mathcal{F}}$ \Comment{projected label embeddings}

\For{$t \gets 1$ \textbf{to} $T$}
  \State \textit{\# Phase 2: Cyclic loss weighting}
  \State $(\lambda_h,\lambda_i,\lambda_c) \gets \Call{CyclicWeights}{t,T,n_{\mathrm{cyc}},W}$

  \ForAll{mini-batch $\mathcal{B}\subset\mathcal{D}$}
  \State \textit{\# Phase 3: MIL forward and dual-branch conditioning}
  \State $z \gets \Call{MILBackbone}{\mathbf{X}}$ \Comment{aggregate instances to slide embedding}
  \State $(v^c,v^f) \gets \Call{SplitAndCondition}{z}$ \Comment{coarse/fine representations}

  \State \textit{\# Phase 4: Decoder input construction (prefix)}
  \State $p \gets \Call{SamplePrompt}{}$ \Comment{template prompt}
  \State $e_{\mathrm{ctx}}^c \gets \Call{Mean}{b^c}$
  \Comment{coarse context token}
  \State $e_{\mathrm{ctx}}^f \gets \Call{Mean}{b^f}$
  \Comment{fine context token}
  \State $E^c \gets \Call{BuildPrefix}{v^c,\langle\mathrm{|img|}\rangle,e_{\mathrm{ctx}}^c,p}$
  \State $E^f \gets \Call{BuildPrefix}{v^f,\langle\mathrm{|img|}\rangle,e_{\mathrm{ctx}}^f,p}$
  \Statex \Comment{$E=\big[\text{img feat},\texttt{<|img|>},\text{context},\text{prompt}\big]$}

  \State \textit{\# Phase 5: Multi-objective training}
  \State $(\hat y^c,\hat y^f) \gets \Call{DualHeadDecode}{E^c,E^f}$ \Comment{shared trunk + two heads}
  \State $\mathcal{L}_{GEN} \gets \Call{CrossEntropy}{\hat y^c,\hat y^f,y^c,y^f}$ \Comment{generation loss}
  \State $\mathcal{L}_{HIER} \gets \Call{TaxonomicConsistency}{t^c,t^f,\mathcal{H}}$ \Comment{HAL + SML}
  \State $\mathcal{L}_{ITAL} \gets \Call{ImageTextAlign}{v^c,v^f,t^c,t^f,y^c,y^f}$
  \State $\mathcal{L}_{CL} \gets \Call{SupConLoss}{v^c,v^f,y^c,y^f}$

  \State \textit{\# Phase 6: Unified optimization}
  \State $\mathcal{L} \gets \mathcal{L}_{GEN} + \lambda_h \mathcal{L}_{HIER} + \lambda_i \mathcal{L}_{ITAL} + \lambda_c \mathcal{L}_{CL}$
  \State $\theta \gets \Call{Update}{\theta, \nabla_{\theta}\mathcal{L}}$
\EndFor
\EndFor

\end{algorithmic}
\end{algorithm}
\FloatBarrier

\section{Prompt Templates}
\label{app:prompts}

We use 10 base prompt templates, which are randomly selected during training. For multi-dataset training, the organ term is substituted by the dataset domain: \textit{gastric} $\rightarrow$ \textit{breast} for BRACS and \textit{gastric} $\rightarrow$ \textit{prostate} for PANDA. At inference time, we use the first template as the default prompt. The prompt templates are provided in Table~\ref{tab:prompt_templates}.

\begin{table}[t]
\centering
\caption{Prompt templates used for training/inference.}
\label{tab:prompt_templates}
\scriptsize
\begin{tabular}{ll}
\toprule
 Setting&Prompt template \\
\midrule
 Training&The observed condition in the gastric image is: \\
 &The diagnosis for the gastric image shows signs of: \\
 &The medical findings for this image indicate: \\
 &This gastric image reveals the presence of: \\
 &The condition detected in the gastric image is: \\
 &The histopathological observation in this image suggests: \\
 &The pathology result from this gastric image describes: \\
 &The condition diagnosed from the image analysis is: \\
 &The elements detected in the gastric region of the image are: \\
 &The visible medical observation in the image corresponds to: \\
\midrule
 Inference&The observed condition in the gastric image is: \\
\bottomrule
\end{tabular}
\end{table}

\section{Datasets}
\label{app:dataset}
We evaluate TaxoMIL on three WSI benchmarks spanning distinct organs and hierarchical diagnostic protocols. Each dataset provides a two-level taxonomy with a coarse-level category and a fine-level subtype or grade.
GastWSI is a private dataset containing 7,228 gastric WSIs with 4 coarse-level categories and 23 fine-level diagnoses. We adopt a $7:1.5:1.5$ train/val/test split to ensure sufficient samples for rare fine-level diagnoses.
BRACS \cite{BRACS} is a public breast pathology benchmark with 3 coarse-level categories and 7 fine-level labels across 545 WSIs with an $8:1:1$ train/val/test split.
PANDA \cite{PANDA} is a public prostate dataset that comprises 10,614 WSIs annotated with ISUP grading. \rev{ISUP grading is a modern, simplified prostate cancer classification scheme that is standard in digital pathology. We retain all 10,614 WSIs without excluding ambiguous cases, a comparatively challenging setting that contributes to the lower absolute performance observed across all models on PANDA.} We adopt a $8:1:1$ train/val/test split and construct a two-level hierarchy by grouping ISUP grades into three risk strata at the coarse-level and using the original ISUP grades as fine-level labels. The specific coarse- and fine-level labels are provided in Table~\ref{tab:data_hierarchy}.

\begin{table}[h]
\centering
\caption{Label hierarchies.}
\label{tab:data_hierarchy}
\scriptsize
\begin{tabular}{lll}
\toprule
Dataset & Coarse label & Fine label \\
\midrule
GastWSI & benign tumor& fundic gland polyp \\
&  & hyperplastic polyp \\
&  & inflammatory polyp \\
&  & xanthoma \\
&  & gastritis cystica polyposa \\
&  & granulation tissue type polyp \\
\cline{2-3}
& cancer& adenocarcinoma, moderately differentiated \\
&  & adenocarcinoma, poorly cohesive carcinoma \\
&  & adenocarcinoma, well differentiated \\
&  & malignant lymphoma \\
&  & NOS\\
&  & neuroendocrine tumor \\
&  & squamous cell carcinoma \\
\cline{2-3}
& dysplasia& low grade \\
&  & high grade \\
&  & indefinite \\
\cline{2-3}
& inflammatory disease& chronic inflammation, mild \\
&  & chronic inflammation, moderate \\
&  & chronic inflammation, marked \\
&  & erosion \\
&  & ulceration \\
&  & intestinal metaplasia \\
&  & neutrophilic activity \\
\midrule
BRACS& benign& normal \\
&  & pathological benign \\
&  & usual ductal hyperplasia \\
\cline{2-3}
& atypical& atypical ductal hyperplasia \\
&  & flat epithelial atypia \\
\cline{2-3}
& malignant& ductal carcinoma in situ \\
&  & invasive carcinoma \\
\midrule
PANDA& normal and low risk& normal \\
&  & low risk, ISUP grade 1 \\
\cline{2-3}
& intermediate risk& intermediate favourable risk, ISUP grade 2 \\
&  & intermediate unfavourable risk, ISUP grade 3 \\
\cline{2-3}
& high risk& high risk, ISUP grade 4 \\
&  & the highest risk, ISUP grade 5 \\
\bottomrule
\end{tabular}
\end{table}

\section{Implementation Details}
\label{app:implementation}
All WSIs are processed at $20\times$ magnification. We use a patch size of $512 \times 512$ pixels for GastWSI and BRACS, and $256 \times 256$ pixels for PANDA. We extract patch embeddings using the frozen UNI\cite{UNI} foundation model as the feature extractor $\mathcal{E}$. 
The shared autoregressive text decoder backbone $\mathcal{T}_{shared}$ is initialized from pretrained GPT-2 small (124M)\cite{GPT-2}, consisting of 12 decoder layers.
We use the standard GPT-2 tokenizer as $\texttt{TOK}$ for all text processing.
For text generation, greedy decoding is implemented with maximum generation lengths of $L^{(c)}=5$ and $L^{(f)}=15$ tokens for coarse- and fine-level predictions, respectively.
The patch- and slide-level embedding dimension is $D = 1024$, and the decoder hidden state size is $H = 768$. 

The model is trained up to 100 epochs with early stopping based on validation performance. We use AdamW with weight decay $0.01$ and a cosine annealing learning-rate scheduler. The learning rate is set to $1\times10^{-5}$ for GastWSI and $1\times10^{-4}$ for BRACS and PANDA, with a batch size of 64.
During training, we set the temperature $\tau=0.07$ for both $\mathcal{L}_{ITAL}$ and $\mathcal{L}_{CL}$, and we apply a margin $m=1.5$ for $\mathcal{L}_{SML}$. For the cyclic loss scheduling, we set the number of cycles to $n_{\text{cycles}}=3$ and define the maximum loss weights as $w_{hier}^{\max}=0.3$, $w_{ital}^{\max}=0.3$, and $w_{cl}^{\max}=0.1$ in all experiments.
All experiments are conducted on an NVIDIA A6000 GPU.

\section{Additional Ablation Analyses}
\label{app:additional_analyses}

\begin{table}[htbp]
\centering
\scriptsize
\setlength{\tabcolsep}{3.5pt}
\renewcommand{\arraystretch}{1.0}
\caption{Performance comparison for TaxoMIL with classification heads and different LLM decoders.}
\label{tab:classifier_decoder}
\begin{tabular}{lllccc ccc ccc}
\toprule
& \multirow{2}{*}{Dataset} & \multirow{2}{*}{Model}
& \multicolumn{3}{c}{Holistic}
& \multicolumn{3}{c}{Coarse}
& \multicolumn{3}{c}{Fine} \\
\cmidrule(lr){4-6}\cmidrule(lr){7-9}\cmidrule(lr){10-12}
& & & ACC& W-F1& $\kappa$& ACC& W-F1& $\kappa$& ACC& W-F1& $\kappa$\\
\midrule
\multirow{6}{*}{\parbox{1.1cm}{Classifier Head}} & \multirow{2}{*}{GastWSI}
& TaxoMIL-Cls & 63.81 & .623 & .595 & 88.00 & .881 & .839 & 65.95 & .627 & .617 \\
& & TaxoMIL     & 64.55 & .618 & .621 & 88.56 & .886 & .847 & 65.86 & .618 & .615 \\
\cmidrule{2-12}
& \multirow{2}{*}{BRACS}
& TaxoMIL-Cls & 72.22 & .726 & .657 & 87.04 & .871 & .778 & 74.07 & .735 & .677 \\
& & TaxoMIL     & 75.78 & .766 & .697 & 90.74 & .909 & .844 & 77.78 & .763 & .718 \\
\cmidrule{2-12}
& \multirow{2}{*}{PANDA}
& TaxoMIL-Cls & 52.35 & .524 & .394 & 70.54 & .682 & .470 & 53.66 & .528 & .409 \\
& & TaxoMIL     & 54.60 & .535 & .413 & 71.29 & .689 & .479 & 55.07 & .536 & .417 \\
\midrule
\multirow{9}{*}{\parbox{1.1cm}{LLM \quad Decoder}} & \multirow{3}{*}{GastWSI}
& BioGPT  & 62.23 & .614 & .580 & 88.56 & .886 & .847 & 63.72 & .613 & .594 \\
& & GPT2-M  & 62.79 & .600 & .618 & 87.35 & .873 & .831 & 63.91 & .602 & .594 \\
& & GPT2-S  & 64.55 & .618 & .621 & 88.56 & .886 & .847 & 65.86 & .618 & .615 \\
\cmidrule{2-12}
& \multirow{3}{*}{BRACS}
& BioGPT  & 72.22 & .733 & .659 & 87.04 & .877 & .786 & 72.22 & .728 & .659 \\
& & GPT2-M  & 66.67& .696& .667& 83.33& .847& .730& 74.07& .734& .676\\
& & GPT2-S   & 75.78 & .766 & .697 & 90.74 & .909 & .844 & 77.78 & .763 & .718 \\
\cmidrule{2-12}
& \multirow{3}{*}{PANDA}
& BioGPT  & 51.31 & .509 & .379 & 71.01 & .682 & .470 & 52.35 & .511 & .388 \\
& & GPT2-M  & 52.63 & .522 & .405 & 70.64 & .680 & .467 & 53.56 & .526 & .404 \\
& & GPT2-S    & 54.60 & .535 & .413 & 71.29 & .689 & .479 & 55.07 & .536 & .417 \\
\bottomrule
\end{tabular}
\end{table}

\rev{\textbf{Non-generative control.}
To isolate the contribution of taxonomy-constrained text generation, we compare TaxoMIL with TaxoMIL-Cls, a variant that replaces the autoregressive decoder with classification heads while keeping the same backbone and taxonomy-aware objectives. As shown in Table~\ref{tab:classifier_decoder}, the generative formulation consistently improves holistic performance across all three datasets, and TaxoMIL-Cls alone still surpasses most baselines, confirming that the core architecture and taxonomy-aware supervision are effective independently of the generation head.}

\noindent\textbf{Decoder choice.}
\rev{As our outputs are short, closed-taxonomy label strings, we adopt GPT-2 small (124M) as a lightweight decoder. We compare it against a domain-specific decoder (BioGPT\cite{biogpt}, 347M) and a larger general decoder (GPT-2 medium, 355M) in Table~\ref{tab:classifier_decoder}. Both alternatives are inferior to GPT-2 small, with GPT-2 medium showing notably lower coarse-level performance on BRACS, likely due to the limited dataset size.}

\begin{table}[htbp]
\centering
\scriptsize
\setlength{\tabcolsep}{6pt}
\caption{Parent--child violation rate (\%) comparison across models.}
\label{tab:pcvr}
\begin{tabular}{lcccc}
\toprule
Model & GastWSI & BRACS & PANDA & Avg. \\
\midrule
TransMIL     & 5.21  & 11.11 & 3.56 & 6.63 \\
CLAM-SB      & 4.56  & 5.56  & 2.72 & 4.28 \\
ABMIL        & 6.23  & 3.70  & 5.44 & 5.12 \\
MambaMIL     & 7.91  & 3.70  & 4.50 & 5.37 \\
S4MIL        & 4.56  & 5.56  & 3.75 & 4.62 \\
HMIL         & 13.67 & 14.81 & 4.32 & 10.93 \\
Chang et al. & 9.76  & 1.85  & 3.75 & 5.12 \\
ViLa-MIL     & 4.84  & 1.85  & 3.19 & 3.29 \\
HiClass      & 7.26  & 7.41  & 3.00 & 5.89 \\
WSI-VQA      & 5.76  & 3.70  & 1.22 & 3.56 \\
SlideChat    & 8.09  & 7.69  & 2.06 & 5.95 \\
TaxoMIL      & 5.30  & 5.56  & 1.41 & 4.09 \\
\bottomrule
\end{tabular}
\end{table}

\noindent\textbf{Hierarchical consistency.}
\rev{In this closed-taxonomy setting, all generation-based methods yield a 0\% invalid-label rate under the exact-match protocol. To further assess consistency, we report the parent--child violation rate (PCVR) in Table~\ref{tab:pcvr}, defined as the proportion of predictions whose fine-level label is inconsistent with the predicted coarse-level label. CAMP is excluded as its paired output structurally prevents violations. TaxoMIL maintains a competitive PCVR while achieving the highest holistic accuracy, indicating a favorable accuracy--consistency trade-off.}

\section{Computational Efficiency}
\label{app:computational_efficiency}

\begin{table}[htbp]
\centering
\scriptsize
\setlength{\tabcolsep}{6pt}
\caption{Computational efficiency comparison across models.}
\label{tab:efficiency}
\begin{tabular}{lcccc}
\toprule
Model & \makecell{Trainable\\Params (M)} & \makecell{Latency\\(ms)} & \makecell{GPU\\(MB)} & GFLOPs \\
\midrule
TransMIL     & 2.69    & 3.16   & 99.37   & 24.99 \\
CLAM-SB      & 0.83    & 0.46   & 20.12   & 1.58 \\
ABMIL        & 0.60    & 0.32   & 18.24   & 1.18 \\
MambaMIL     & 4.00    & 1.02   & 55.32   & 7.87 \\
S4MIL        & 1.10    & 0.80   & 163.83  & 1.07 \\
HMIL         & 1.35    & 1.10   & 27.33   & 6.45 \\
Chang et al. & 0.60    & 2.49   & 18.22   & 1.18 \\
ViLa-MIL     & 47.21   & 16.55  & 267.35  & 166.23 \\
HiClass      & 4.33    & 0.94   & 33.47   & 1.58 \\
WSI-VQA      & 23.96   & 288.92 & 235.56  & 701.39 \\
SlideChat    & 3117.45 & 97.33  & 6671.69 & 6352.55 \\
CAMP         & 55.34   & 204.77 & 1834.67 & 84.37 \\
TaxoMIL      & 218.77  & 114.37 & 897.44  & 14.08 \\
\bottomrule
\end{tabular}
\end{table}

\noindent{Table~\ref{tab:efficiency} compares trainable parameters, latency, GPU memory, and GFLOPs. The text decoder introduces overhead over standard MIL aggregators, yet TaxoMIL requires far fewer GFLOPs and achieves latency comparable to VLM baselines.}

\end{document}